\def\year{2022}\relax
\documentclass[letterpaper]{article} 
\usepackage{aaai22}  
\usepackage{times}  
\usepackage{helvet}  
\usepackage{courier}  
\usepackage[hyphens]{url}  
\usepackage{graphicx} 
\usepackage{natbib}  
\usepackage{caption} 
\usepackage{multirow}
\usepackage[utf8]{inputenc}
\usepackage[T1]{fontenc}
\usepackage{mathtools}
\usepackage{natbib}
\usepackage{hhline}
\usepackage{amsfonts}
\usepackage[bb=boondox]{mathalfa}
\DeclareCaptionStyle{ruled}{labelfont=normalfont,labelsep=colon,strut=off} 
\usepackage{algorithm}
\usepackage{algorithmic}

\usepackage{newfloat}
\usepackage{listings}
\floatstyle{ruled}
\newfloat{listing}{tb}{lst}{}
\floatname{listing}{Listing}
\usepackage{array}
\newcolumntype{C}{>{\centering\arraybackslash}m{4em}}

\title{CABACE: Injecting Character Sequence Information and Domain Knowledge for Enhanced Acronym and Long-Form Extraction}
\author {
    Nithish Kannen, 
    Divyanshu Sheth, 
    Abhranil Chandra, 
    Shubhraneel Pal 
}
\affiliations {
    Indian Institute of Technology Kharagpur, West Bengal, India\\
    
    \{nithishkannen, shethdivyanshu2000, abhranil.chandra, shubhraneel\}@iitkgp.ac.in
}

\usepackage{bibentry}
\usepackage{adjustbox}

\begin{document}

\maketitle

\section{Abstract}
Acronyms and long-forms are commonly found in research documents, more so in documents from scientific and legal domains. Many acronyms used in such documents are domain-specific, and are very rarely found in normal text corpora. Owing to this, transformer-based NLP models often detect OOV (Out of Vocabulary) for acronym tokens, especially for non-English languages, and their performance suffers while linking acronyms to their long forms during extraction. Moreover, pre-trained transformer models like BERT are not specialized to handle scientific and legal documents. With these points being the overarching motivation behind this work, we propose a novel framework \textbf{CABACE}: \textbf{C}haracter-\textbf{A}ware \textbf{B}ERT for \textbf{AC}ronym \textbf{E}xtraction, which takes into account character sequences in text, and is adapted to scientific and legal domains by masked language modelling. We further use an objective with an augmented loss function, adding max loss and mask loss terms to the standard cross-entropy loss for training CABACE. We further leverage pseudo labelling and adversarial data generation to improve the generalizability of the framework. Experimental results prove the superiority of the proposed framework in comparison to various baselines. Additionally, we show that the proposed framework is better suited than baseline models for zero-shot generalization to non-English languages, thus reinforcing the effectiveness of our approach. Our team BacKGProp secured the highest scores on the French dataset, second-highest on Danish and Vietnamese, and third-highest in English-Legal dataset on the global leaderboard for the acronym extraction (AE) shared task at SDU AAAI-22.



\section{Introduction}
\noindent Acronyms are short forms used to represent a longer sequence of words in documents, for brevity. Most commonly, they are generated by joining the starting letter/letters of each word in their long-form. Such shorthand notations help writers save space and avoid redundant mentions of long-forms in text, especially in scientific papers, which have a page/word limit. A large number of acronyms that occur in scientific and legal documents are domain-specific and are almost never found in common text corpora. As a result, these acronyms often go into the (OOV) -- Out of Vocabulary category of NLP models, especially for languages other than English. However, these acronyms quite often form the subject of sentences and play a crucial role in document understanding or text analytics in the scientific domain \cite{veyseh2021acronym}.

\begin{figure}[t!]
  \centering
  \includegraphics[keepaspectratio,width=0.45\textwidth]{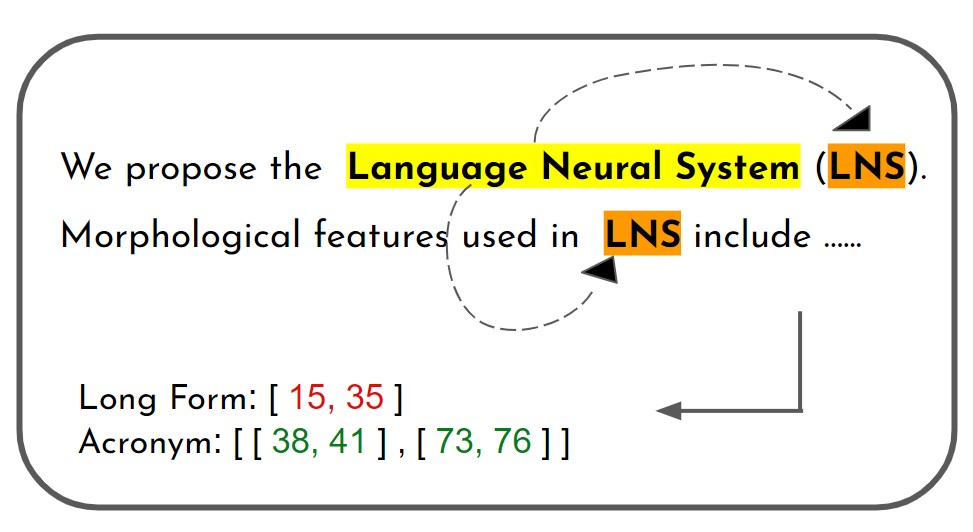}
  \caption{ \footnotesize An example sentence depicting {long-forms} and acronyms appearing together. The goal of the task is to extract the character span indices of these occurrences from documents.}
  
  \label{fig:intro} 
\end{figure}

In order to avoid misconceptions among readers, most documents provide the long form of rare acronyms at least at the time of their first mention. As a result, it is important that NLP systems built for scientific document understanding take this into account and look within the documents themselves to understand acronyms and their provenance. For example, in Fig \ref{fig:intro}, the long-form, “Language Neural System” is introduced to the reader, and it is referred to by an acronym, “LNS” thereafter. Systems that can identify acronyms within the passage, and can further link these to their corresponding long-forms, if present, would enable models to comprehend scientific documents much better. The ease of availability of scientific documents on arXiv\footnote{https://arxiv.org/} and other open-access archives have led to increased research interest on scientific documents \cite{veyseh2021acronym, Timmapathini2021ProbingTS, li2021unsupervised}.





Although acronym extraction has been studied in the past, the datasets and models used have mostly been limited to the biomedical domain, overlooking challenges in other domains such as science or legal. With this very objective in mind \cite{veyseh2021acronym} made the largest manually annotated acronym identification and disambiguation dataset publicly available at the SDU Workshop at AAAI-21, which saw systems improving over the baselines. The superiority of human performance in comparison to the best system proposed was highlighted, validating the scope of future research. At SDU-AAAI-22, the acronym extraction task is extended to 6 different languages, namely English, French, Spanish, Danish, Vietnamese and Persian, making it the first publicly available multilingual benchmark for acronym extraction on scientific documents \cite{veyseh-et-al-2022-MACRONYM}
. Given that multilingual NLP/low-resource NLP is starting to pick up pace, such a multilingual benchmark further enables us to test the efficacy of systems on a diverse set of languages.

In this paper, we elucidate our unified framework \textbf{CABACE}: \textbf{C}haracter-\textbf{A}ware \textbf{B}ERT for \textbf{AC}ronym \textbf{E}xtraction, adapted for 5 languages (all provided in the shared task, except Persian), modelling the acronym extraction (AE) task as a sequence labelling problem. As mentioned earlier, many of the acronyms in scientific text rarely occur in the vocabulary of BERT/Multilingual BERT (mBERT) \cite{devlin2019bert}. Towards this end, we utilise character sequence information of tokens, and aggregate individual character embeddings using a convolutional neural network (CNN) layer, followed by max-pooling. In this way, we inject character sequence information into our model along with mBERT token embeddings. We leverage domain-specific language modelling to enrich mBERT embeddings with domain knowledge, and further improve model generalizability using pseudo labelling and adversarial data augmentation. Additionally, we perform masking of tokens with positive labels (acronyms or long-forms) to encourage CABACE to pay higher attention to the context, and use an objective with an augmented loss function, adding max-loss and mask-loss terms to the standard cross entropy loss. Finally, we also evaluate and provide our system performances on zero-shot transfer from English to the French, Spanish, Danish and Vietnamese datasets. Our proposed model achieved the highest score on the French dataset, and the second highest on Danish and Vietnamese, third highest on Legal English, and fourth on Spanish and Scientific English on the AE Shared Task \cite{veyseh-et-al-2022-Multilingual}\footnote{https://sites.google.com/view/sdu-aaai22} at SDU AAAI-22. We make our code publicly available\footnote{https://github.com/nitkannen/BacKGProp-AAAI-22}.

To summarize, the key contributions of this work are:

\begin{itemize}
\item We propose a novel framework CABACE that leverages mBERT and aggregates character embeddings using CNN and max-pooling to form character-aware token embeddings, which are used along with mBERT token representations. We also inject domain knowledge via masked language modelling performed on scraped data, which is shown to improve AE performance.
\item We further use an objective with an augmented loss function, adding max and mask loss terms in addition to standard cross-entropy loss, that results in improved performance when paired with CABACE. We use pseudo labelling and adversarial data generation to improve model generalizability.
\item We test the zero-shot generalization efficacy of CABACE across languages and show that it performs better than vanilla mBERT. To the best of our knowledge, this is the first work reporting zero-shot efficacy of acronym extraction systems on a multilingual benchmark. 
\item We perform extensive experiments on the AE task of SDU-22, achieving state-of-the-art results in both normal and zero-shot settings, demonstrating the effectiveness of our approach.
\end{itemize}

\begin{table*}[!t]
\centering
\small
\renewcommand{\arraystretch}{1.5}

\begin{tabular}{*{10}{C}}
\hhline{----------}
\textbf{Dataset}  & \textbf{Train Set Size} & \textbf{Dev Set Size} & \textbf{Test Set Size} & \textbf{Avg. Word Length} & \textbf{Avg. num. of Ac.} & \textbf{Avg. num. of LF} & \textbf{\ Num. with Ac. + LF} & \textbf{\ Num. with only Ac.} & \textbf{\ Num. without Ac./LF}\\ \hline
\hhline{----------}
Danish & 3082 & 385 & 386 & 64.04 & 2.04 & 0.69 & 2215 & 915 & 336 \\ \hline
Eng-Leg. & 3564 & 445 & 446 & 66.30 & 2.68 & 1.48 & 3913 & 86 & 10 \\ \hline
Eng-Sci. & 3980 & 497 & 498 & 29.56 & 1.93 & 1.44 & 4457 & 19 & 1\\ \hline
French  & 7783 & 973 & 973 & 80.47 & 2.79 & 1.74 & 8585 & 161 & 10\\ \hline
Spanish & 5928 & 741 & 741 & 85.39 & 2.18 & 1.57 & 6649 & 16 & 4 \\ \hline
Vietnamese & 1274 & 159 & 160 & 33.80 & 1.05 & 0.05 & 215 & 807 & 410\\ \hline

\end{tabular}
\caption{Key statistics of the released datasets for the Danish, English, French, Spanish and Vietnamese languages, which we evaluate our systems over. The English data has two splits, legal and scientific. We report the average word length and the average number of acronyms and long-forms for a datapoint in each dataset. We also report the number of datapoints containing both acronyms and long-forms, only acronyms, and those with neither acronyms nor long-forms on the train+dev combined set.}
\label{Table1}
\end{table*}

\section{Related Work}
Models trained for acronym extraction tasks, like most NLP approaches, can be divided into three major categories: 1) rule-based methods \cite{Taghva1999, Schwartz03asimple}, 2) machine learning methods that use text-based features \cite{Kuo2009, HarrisChristopherG2019MWMv}, 3) deep learning methods. 
\cite{singh2021scidr} used BERT and SciBERT with BIO less tagging and blending in an ensemble framework for acronym identification. \cite{Li2021SystemsAS} experimented with multi-task learning, feature engineering and CRF, and found that feature-based methods handled the task well. With the advent of large pretrained language models such as BERT \cite{devlin2019bert}, which is a bidirectional multi-layer transformer encoder that achieved state-of-the-art results on a number of benchmarks at the time, most NLP tasks have seen use of transformers-based architectures. \cite{zhu2021atbert} used the BERT model with adversarial samples that were created by perturbing existing inputs such that the loss of the model would increase on those samples, to improve the robustness of BERT. Their system was the winning solution to the acronym identification shared task at SDU AAAI-21.

\section{Dataset Statistics and Task Description}

The organizers of SDU-22 provide acronym extraction (AE) datasets for the shared task in 6 languages, namely English, Spanish, French, Danish, Vietnamese, and Persian 
There are two distinct splits to the English data, one for texts from the scientific English domain and one for texts from the legal English domain. All datasets are provided in the form of json files, with each individual datapoint having raw text, an ID, and a list of ground truth acronyms and long-forms present in the raw text. Ground truth acronym and long-forms for each datapoint are provided as separate lists, with elements in the list being of the form -- \textit{[starting character index, ending character index]} for each acronym/long-form in the text. The objective of the acronym extraction task is to extract character spans for each identified acronym and long-form in given sentences. Table \ref{Table1} lists key statistics of the datasets provided in 5 languages.

\section{Methodology}

Given the input sentence, our objective is to jointly predict the character span of acronyms and long forms present in the text. Towards this goal, we first explain how we formulate the problem. Then we go into the details of the proposed CABACE framework, following which we describe techniques we adopted for robustness and to reduce overfitting, i.e., pseudo-labelling and adversarial data generation.

\subsection{Problem Formulation}

With the given dataset being annotated with character span indices of acronyms and long-forms, we preprocess the dataset to convert it into sequence tags to model it as a sequence labelling problem. Specifically, we use the BIO tagging scheme for labelling all tokens in the target sequence. We have 5 possible label tags:\ 0) O-None, 1) B-Acronym, 2) I-Acronym, 3) B-Longform, 4) I-Longform, where 'B' stands for begin, and 'I' stands for inside. The first token for an acronym/long-form would be given a B-label, and the rest would be given I-labels. For all our experiments, we use mBERT that uses the WordPiece tokenizer, whose property we leverage for converting character spans to BIO-tags as explained in Algorithm \ref{alg:idx2BIO}. 

The evaluation script provided by the task organizers uses character spans to calculate eval. metrics, so we convert the sequence tags back to character span index after model prediction. For this, we leverage \textit{offset map} returned by BertTokenizer. More specifically, for each positive prediction by our model (either acronym or long form), we use the token's span to finally compute the character span range of predicted acronyms/long forms.

\begin{figure*}[!t]
    \centering
    \includegraphics[width=2\columnwidth]{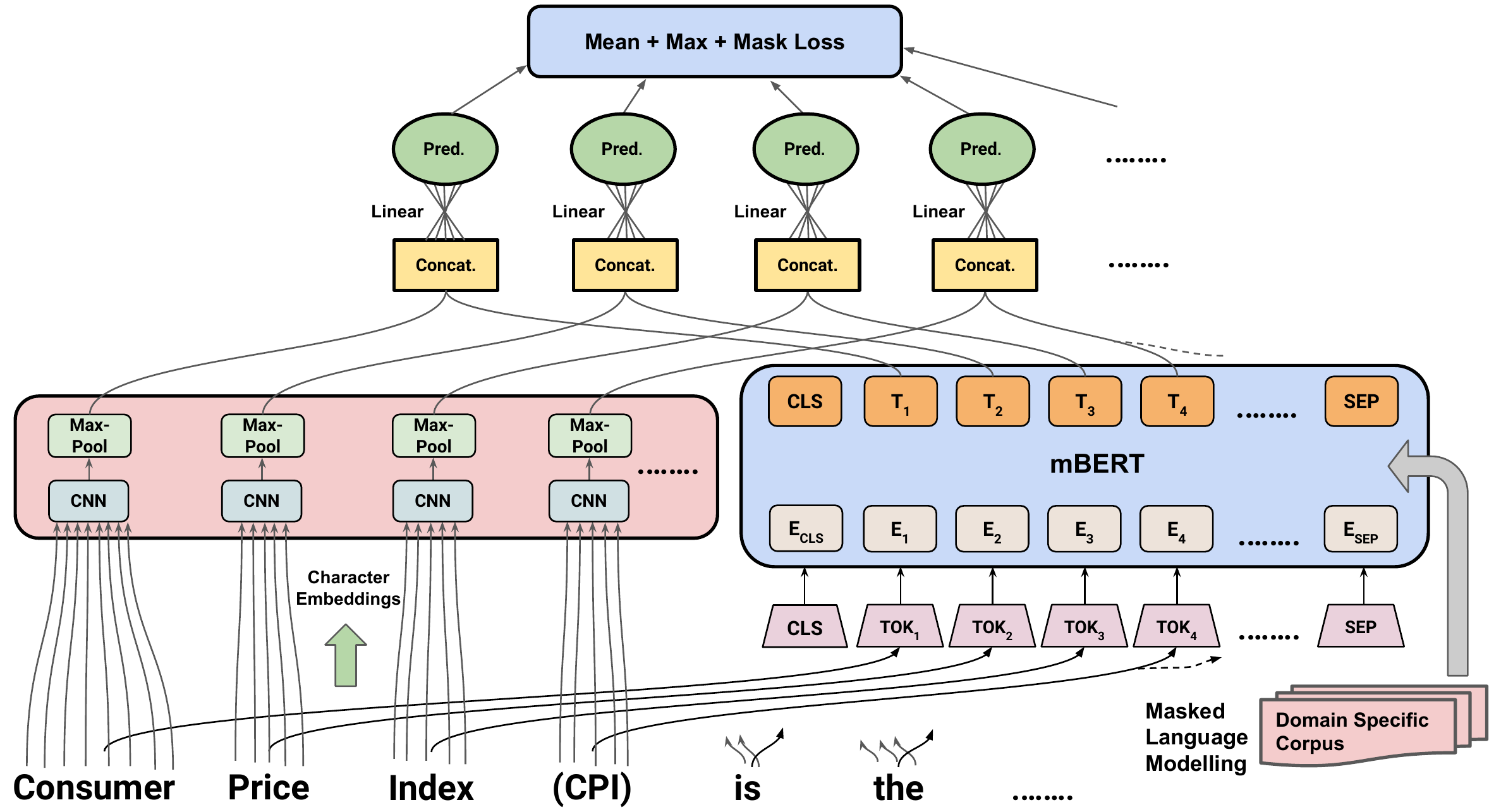}
    \caption{The CABACE Architecture. Input tokens are passed to mBERT (right) and to the CNN \& max-pooling layers (left) character-by-character (using character embeddings). The resulting outputs from both are concatenated and passed through a prediction layer (linear + softmax) before computing the augmented loss function. Note that the token '(CPI)' gets split into sub-words by mBERT tokenizer.}
    \label{fig:overall}
\end{figure*}


\begin{algorithm}[tb]
\caption{Conversion of character spans to BIO tags}
\label{alg:idx2BIO}
\textbf{Input}: $Sentence$, $Tokenizer$, $AcronymList$\\
\textbf{{Output}: $target$ }      
\begin{algorithmic}[1] 
\STATE $tokens$ = $Tokenizer$($Sentence$)
\STATE $target$ = [0] * len($tokens$)

\FOR {each $ac\_text$  in $Acronym\_List$}


\STATE $ac\_tokens$ = $Tokenizer(ac\_text)$

\FOR {each i in $range(len(tokens))$}
\IF {$tokens$[i : i + len($ac\_tokens$)] = $ac\_tokens$}
\STATE    {FillBIO}(i, i + len($ac\_tokens$), $target$)
\ENDIF
\ENDFOR
\ENDFOR

\STATE \textbf{return} target

\end{algorithmic}
\end{algorithm}

\subsection{The CABACE Framework} We now present \textbf{CABACE}, \textbf{C}haracter-\textbf{A}ware \textbf{B}ERT for \textbf{AC}ronym \textbf{E}xtraction.
Our proposed framework takes as input the tokens of the sentence, and classifies each token into one of the 5 possible labels. For this we augment mBERT with the following components as in Fig. \ref{fig:overall}: 1) Masked language modelling, to inject domain specific knowledge into mBERT to make it better equipped to handle scientific documents, 2) Character-aware token embeddings, to provide information about character sequences in each token, 3) Augmented loss and masking, motivated by the presence of rarely used notations as acronyms. We describe these components in great detail in the upcoming sections.

\subsubsection{Domain-Specific Language Modelling:}
To inject scientific domain knowledge into mBERT, which is otherwise not a significant portion of mBERT's pretraining, we perform domain-specific language modeling. For this, we scrape sentences containing acronyms from the web. For scientific English, we scrape data from the arXiv API\footnote{https://arxiv.org/help/api/} and for French, Spanish, and Danish, we scrape data from Wikipedia. Then, from the scraped sentences, we filter out those sentences which contain a probable acronym or long-form. A sentence containing a word with more than 50\% capitalized letters was used as the criteria to detect sentences with a probable acronym and long-form. We also use the acronym identification data from last year's SDU-21. We merge all these datasets for each language with the datasets provided for this shared task. The total number of sentences thus obtained is 25009 for English Scientific, 4455 for English Legal, 19769 for French, 17411 for Spanish, 15196 for Danish and 1593 for Vietnamese. With this data, we perform masked language modeling (MLM), following the same procedure as described in \cite{devlin2019bert}. We start the training from public mBERT checkpoint and train for 6 epochs. The resulting model weights are saved and are an integral part of the CABACE framework. We use these LM checkpoints with CABACE in all our experiments unless mentioned otherwise.

\subsubsection{Character-Aware Token Embeddings:}

A variety of acronyms found in scientific documents often have repeated suffices/prefixes common to them that can be important signals used to detect such acronyms. For example, the acronyms \textit{MIT, IIT and NIT} have common suffices that can be overlooked if WordPiece tokenizer of mBERT\footnote{https://huggingface.co/bert-base-multilingual-cased/blob/main/tokenizer.json} fails to segment out the common suffix 'IT' that serves as an important signal to detect acronyms. Apart from this, characters/sub-tokens that are out of mBERT's vocabulary get mapped to the [UNK] token, where important case-specific or character specific information crucial for the task can be lost. Such cases or more common in low-resource languages like Vietnamese. Motivated by these shortcomings, we propose to additionally inject character-aware token embeddings that are concatenated with mBERT final layer token embeddings before being passed to a prediction layer. 

With reference to Fig. \ref{fig:overall}, we pass the character embeddings of each character of a token to a convolutional neural network (CNN) layer of a fixed filter size. For pre-trained character embeddings, we use the FastText library\footnote{https://fasttext.cc/}, the dimension of the character embeddings being 300. We pad each input token to a constant character length. Hence, for a token of character length 16 (where max pad length is also 16), we would pass a 16 X 300 dimensional vector $\in \mathbb{R}^{16\times300}$ into the CNN. We then use a max-pooling layer that picks out the CNN filter output that contains the highest signal. This way, we get one embedding vector for each token in the input, which likely contains the most important character n-gram signal from the token (where 'n' refers to the CNN filter size). Let \[ \mathcal{D} =  [ tok_1,tok_2, .... tok_{k-1},  tok_k ] \] where $\mathcal{D}$ is a sentence with k tokens. The computation for each token $tok_i$ can be summarized by the following equations, where $\tilde{h_i}$ is passed into a linear layer for classification:

$$\tilde{s_i} = BERT(tok_i)$$
$$\tilde{e_i} = MaxPool(CNN(tok_i))$$
$$\tilde{h_i} = \tilde{s_i}  ||  \tilde{e_i}$$

\subsubsection{Augmented Loss and Masking:} 
Compared to commonly used words, acronyms occur with very low-frequency in the training corpora, and with even lower frequency in the test corpora. As a result, the model must be encouraged to pay higher attention to the context around acronyms to reduce the dependence on the acronym token itself during prediction. Such a feature could be applicable to long forms as well, although with less significance. Towards this end, we randomly mask 10\% of tokens with positive labels (\textit{B-} or \textit{I-} tokens) during the training phase. This encourages the model to rely less on the token itself, and more on the context leading up to an acronym/long form. Inspired by the successful amendment to the loss function in previous works \cite{wu2020enhanced}, we additionally append a max term that adds the maximum loss value across all token labels from a particular example to the standard cross entropy loss. This ensures that the model learns more from wrong predictions with high losses, as opposed to an uniformly weighted loss that doesn't special pay attention to the token with the highest loss. Let \[ \mathcal{L} = [Loss(tok_1), Loss(tok_2), ... , Loss(tok_n) ] \]
where Loss(.) is given by: $Loss(\hat{p}) = -p\ log(\hat{p})$ \\

\noindent Our new objective function contains a weighted addition of max and mask loss terms along with cross entropy.

$$
\begin{aligned}
CrossEntropyLoss & = mean(\mathcal{L}) \\
MaxLoss & = max(\mathcal{L}) \\
MaskLoss & = \sum_{i=[\mathrm{MASK}]}Loss(tok_i)
\end{aligned}
$$

\noindent The max term is weighted by $\lambda_{max}$ and the mask term is weighted by $\lambda_{mask}$ which are hyperparameters. The overall augmented loss is given by the following equation.

$$
\begin{aligned}
Augmented Loss\ = &\ CrossEntropyLoss\ \\
        + &\ \lambda_{max} MaxLoss\ \\+ &\ \lambda_{mask} MaskLoss
\end{aligned}
$$

\subsection{Pseudo-Labelling} Pseudo-labelling \cite{iscen2019label} is where additional data is created by running a trained model on unseen data and using the model's high-confidence predictions on the unseen data as ground-truths, adding them to the training dataset. We train our models from the public mBERT checkpoint on the training dataset of the original SDU-22 dataset. The trained models are used on the unlabeled scraped data to identify acronyms and long-forms in them. We append the datapoints with high-confidence predictions back to our training set. Pseudo-labeling is a very useful semi-supervised data generation technique \cite{cascantebonilla2020curriculum} that helps particularly when working with low-resource datasets.

\subsection{Adversarial Data Generation}
We perform adversarial data generation for the English datasets. Adversarial training is a useful technique that enhances the robustness \cite{goodfellow2015explaining} of models by adding adversarial samples to the training data. An adversarial example is an instance with small, intentional feature perturbations that induces the model to make a false prediction \cite{szegedy2014intriguing}. In the procedure of adversarial training, input samples are first mixed with some small perturbations to generate adversarial samples. The model is then trained with both the original input sample and generated adversarial samples \cite{zhu2021atbert}. We use the embedding function of the augmenter class from TextAttack \cite{morris2020textattack} to generate text by replacing words with neighbors in the counter-fitted embedding space consisting of GloVe vectors, with a constraint to ensure their cosine similarity is at least 0.8. One such example is:\\
\textit{Original Text:} "We conduct a showcase study of dialectal language in online conversational text by investigating African-American English (AAE) on Twitter."\\
\textit{Generated Adversarial Example:} "We conduct a case study of dialectal language in online conversational text by scrutinize African-American English (AAE) on Twitter."

\section{Experiments}

We compare our model with 3 baselines explained in the next section. We then go over the evaluation metrics and implementation details, followed by experiments to test cross-lingual zero-shot efficacy.




\subsection{Baselines}

We compare our proposed model with 3 different baselines. These are explained in the coming sections:

\subsubsection{Rule-Based:}
We report the results obtained by the rule-based baseline provided by the organizers of the shared task\footnote{https://github.com/amirveyseh/AAAI-22-SDU-shared-task-1-AE/blob/main/code/baseline.py}. This system uses hand-picked rules for extracting acronyms and long-forms. Words that have >60\% capitalization are selected as acronyms, and if the initial characters of the preceding words before an acronym can form the acronym, those words are selected as the long-form.

\subsubsection{Vanilla mBERT for Sequence Labelling:}
Inspired by the success of BERT \cite{devlin2019bert} for sequence labelling in acronym identification shared task of SDU-21 \cite{zhu2021atbert}, as well as its ability to effectively aggregate contextual information from texts, we consider it as a baseline in our experiments. We run our inputs through mBERT, and use mBERT's final layer tokens, which are then passed to a linear classifier followed by softmax to generate 1 out of the 5 BIO tags as prediction for that token. Problem formulation and pre-processing is the same as in CABACE. We use Multilingual-BERT (bert-base-multilingual-cased)\footnote{https://huggingface.co/bert-base-multilingual-cased} and refer to this model in our experiments as Vanilla mBERT.

\subsubsection{Seq-to-Seq:} Previous works have reported sequence to sequence approach using the encoder-decoder architecture as an alternative for the sequence tagging scheme \cite{almasian2021bert, yan2021unified}. Following these works, we use a transformer-based generative framework to auto-regressively decode the acronyms and long-forms present in input sentence. For the example given in Fig.\ref{fig:intro}, the ground truth target sequence would follow the template string: "\textit{<Acronyms> LNS <Long-Forms> Language Neural System}", where "\textit{<Acronyms>}" and "\textit{<Long-Forms>}" contain the predicted acronyms and long-forms from the sentence, separated by a comma. We decode the output by searching for occurrences of the predicted acronyms and long-forms and detecting their character spans in the input text. We use mT5 for our experiments \cite{xue2021mt5}.


\subsection{Zero-Shot Transfer to Non-English Languages}
Multilingual models like mBERT have a shared embedding space across languages, which they leverage to learn language-agnostic properties of the task along with language-specific features. Zero-shot transfer is a useful way to test how well multilingual models generalize to unseen languages \cite{choi2021analyzing}. To this end, we conduct experiments by training models on a combined English-Legal + English-Scientific dataset, and testing their performances on the other languages, i.e., Danish, French, Spanish and Vietnamese datasets. We comparatively evaluate Vanilla mBERT and CABACE this way to test their zero-shot efficacy.



\begin{table}
\centering
\small
\renewcommand{\arraystretch}{1.3}
\begin{tabular}{ll}
\hline
\textbf{Hyperparameter}  & \textbf{Value}\\ \hline \hhline{==}
Batch size & 8 \\ \hline
Token character len. & 16\\ \hline
CNN filter size & 4 \\ \hline
$\lambda_{max}$ & 2.0 \\ \hline
$\lambda_{mask}$ & 1.0 \\ \hline
Mask rate & 0.1 \\ \hline
Learning rate & 2e-5 \\ \hline

\end{tabular}
\caption{Hyperparameters used for the CABACE model}
\label{hyperpTable}
\end{table}

\subsection{Evaluation Metrics}
The metrics reported here is identical to the ones provided by the organizers of the shared task\footnote{https://github.com/amirveyseh/AAAI-22-SDU-shared-task-1-AE/blob/main/code/scorer.py}. The metrics used were precision, recall and F1-score for the acronyms and long-forms, both individually and combined together.
The script uses exact match to pair predicted spans with the gold spans.



\subsection{Implementation Details}

All experiments were done using Pytorch 1.10.0+cu11.1 on Google Colab GPUs (NVIDIA Tesla P100-PCIe-16GB). For the mBERT and mT5 checkpoints, we used the HuggingFace transformers 4.12.5 library\footnote{https://huggingface.co/transformers/}. We used the \textit{base-base-multilingual-cased} version for mBERT and the \text{base} version for mT5. HuggingFace datasets 1.15.1 was used to handle dataset processing. For fine-tuning, we used the AdamW optimizer with learning rate 2e-5 along with a learning rate scheduler with warmup, wherein the learning rate decreases from its initial value to 0 after a warmup period of 0 to the initial value. We set gradient clipping to 1.0 to prevent exploding gradients. The maximum sequence length for finetuning mBERT was taken as 512. The maximum sequence length for both input and output in mT5 was taken to be 600. For masked language modelling, we used the same scheme as mentioned in \citet{devlin2019bert}. All results are reported on the development set. Table \ref{hyperpTable} lists the hyperparameters used in our experiments.


\begin{table}[!b]
\centering
\small
\begingroup
\setlength{\tabcolsep}{6pt} 
\renewcommand{\arraystretch}{1.0} 

\begin{tabular}{ccccc}
\hline
\textbf{Datasets} & \textbf{Model} & \textbf{Precision} & \textbf{Recall}& \textbf{F1}  \\  \hhline{=====}

\multirow{4}{4em}{Danish} &
Rule-Based & 0.1000 & 0.0600 & 0.0800 \\&Seq-to-Seq (mT5) & 0.5773 & 0.6821 & 0.6254 \\
& Vanilla mBERT & 0.9285 & 0.9515 & 0.9398 \\ 
&\textbf{CABACE (Ours)}& \textbf{0.9435} & \textbf{0.9572} & \textbf{0.9503}\\ \hline
  
\multirow{4}{4em}{English Legal} &
Rule-Based & 0.3200 & 0.1000 & 0.1600\\&Seq-to-Seq (mT5) & 0.7024 & 0.6398 & 0.6697 \\& Vanilla mBERT & 0.8592 & 0.8727 & 0.8659 \\ 
&\textbf{CABACE (Ours)}& \textbf{0.8593} & \textbf{0.8756}  & \textbf{0.8681}\\ \hline
    
\multirow{4}{4em}{English Scientific} &
Rule-Based & 0.3300 & 0.1500 & 0.2000 \\&Seq-to-Seq (mT5) &0.8000  & 0.7455 &0.7718  \\
& Vanilla mBERT & 0.8108 & 0.8535 & 0.8316 \\ 
&\textbf{CABACE (Ours)}& \textbf{0.8282}& \textbf{0.8876}  & \textbf{0.8509}\\ \hline
    
\multirow{4}{4em}{French} &
Rule-Based & 0.2200 & 0.0600 & 0.1000 \\&Seq-to-Seq (mT5) & 0.7891 & 0.6771 & 0.7288 \\&
Vanilla mBERT & 0.9133 & 0.9168 & 0.9150 \\ 
&\textbf{CABACE (Ours)}& \textbf{0.9387} & \textbf{0.9423}  & \textbf{0.9405}\\ \hline

\multirow{4}{4em}{Spanish} &
Rule-Based & 0.1700 & 0.0700 & 0.1000 \\&Seq-to-Seq(mT5) & 0.7544 & 0.6604 & 0.7043 \\&
Vanilla mBERT & 0.8656 & 0.8770 & 0.8712  \\ 
&\textbf{CABACE (Ours)}& \textbf{0.8842}& \textbf{0.9029} & \textbf{0.8934}\\ \hline
    
\multirow{4}{4em}{Vietnam-ese} &
Rule-Based & 0.8200 & 0.3900 & 0.5300 \\&Seq-to-Seq(mT5) & 0.50 & - & - \\&
Vanilla mBERT & 0.7852 & 0.6589 & 0.7165 \\
&\textbf{CABACE (Ours)} & \textbf{0.9077}& \textbf{0.7839}  & \textbf{0.8413}\\ \hline

\end{tabular}
\endgroup
\caption{Comparison of CABACE with the three baseline models. Reported scores are on the dev set. CABACE is seen to outperform all baseline methods in all datasets.}
\label{mainTable}
\end{table}

\section{Results and Discussion}

Table \ref{mainTable} lists comparative results between CABACE and the Rule-based model, the Seq-to-seq (mT5) and Vanilla mBERT baselines. We find that CABACE improves significantly upon the three baselines in all 6 datasets we evaluate the models upon. Note that the improvement with CABACE in a language like Vietnamese is much more than in English-Legal. This can be attributed to the fact that mBERT classifies many of the tokens in Vietnamese into [UNK], consequently losing crucial information. Injecting character embeddings for these tokens enables CABACE to perform much better than Vanilla mBERT. Vanilla mBERT performs better than the Seq-to-seq model, which in turn ups the rule-based model's performance, in all datasets.

Table \ref{fine-grained} shows fine-grained results of the CABACE architecture on the 6 datasets -- precision, recall and F1-scores for acronyms and long-forms are provided separately. We observe that CABACE is able to perform very well on extracting acronyms, but loses some points while detecting long-forms, which could be due to lack of components in the architecture that are specialized to focus on the acronym-long form interactions during extraction. Recall scores are seen to be always higher than precision scores, except on the Vietnamese dataset.

A comparison of Vanilla mBERT vs CABACE for zero-shot performance can be found in Figure 3. We find that CABACE outperforms Vanilla mBERT in 3 out of 4 languages for cross-lingual zero-shot transfer. Surprisingly, the zero-shot performance in Spanish using both models isn't much less than performances of the models when trained on the Spanish dataset. However, languages like Vietnamese and Danish see significant zero-shot performance drops compared to in-dataset performance. Possible reasons could be due to lexicons and grammar in these languages being relatively distant to those in English, and relative similarity of Spanish and English.

\begin{table}[!t]
\centering
\small
\begingroup
\setlength{\tabcolsep}{6pt} 
\renewcommand{\arraystretch}{1.0} 

\begin{tabular}{ccccc}
\hline
\textbf{Datasets} & \textbf{Fine-grained} & \textbf{Precision} & \textbf{Recall}& \textbf{F1}  \\ \hhline{=====}

\multirow{4}{4em}{Danish} &
Acronyms & 0.9637 & 0.9809 & 0.9722 \\
&  Long-Forms & 0.9234 & 0.9336 & 0.9284 \\
&{Combined}& {0.9435} & {0.9572}  & {0.9503}\\ \hline

\multirow{4}{4em}{English Legal} &
Acronyms & 0.8870 & 0.9126 & 0.8996 \\
&  Long-Forms & 0.8190 & 0.8386 & 0.8287 \\
&{Combined}& {0.8593} & {0.8756}  & {0.8681}\\ \hline

\multirow{4}{4em}{English Scientific} &
Acronyms & 0.9038 & 0.9299 & 0.9167 \\
&  Long-Forms & 0.7968 & 0.8222 & 0.8093 \\
& {Combined}& {0.8282} & {0.8876}  & {0.8509}\\ \hline

\multirow{4}{4em}{French} &
Acronyms & 0.9599 & 0.9491 & 0.9541 \\
&  Long-Forms & 0.9208 & 0.9269 & 0.9173 \\
&{Combined}& {0.9387} & {0.9423}  & {0.9405}\\ \hline

\multirow{4}{4em}{Spanish} &
Acronyms & 0.9308 & 0.9447 & 0.9377 \\
&  Long-Forms & 0.8376 & 0.8610 & 0.8491 \\
&{Combined}& {0.8842} & {0.9029}  & {0.8934}\\ \hline

\multirow{4}{4em}{Vietnam-ese} &
Acronyms & 0.9821 & 0.9429 & 0.9621 \\
&  Long-Forms & 0.8333 & 0.6250 & 0.7143 \\
&{Combined}& {0.9077} & {0.7839}  & {0.8413}\\ \hline

\end{tabular}
\endgroup
\caption{Fine-grained performance metrics using CABACE Combined depicts the overall score, while Acronyms and Long-Forms extraction metrics are individually reported.}
\label{fine-grained}
\end{table}

\subsection{Ablation Study on CABACE}
We perform an ablation study to better interpret how individual components contribute to performance improvements. The steps we perform include comparing our full model (CABACE), CABACE without the max loss component, CABACE without the mask loss component, CABACE without the character embeddings, and CABACE without language modelling. We perform the ablation study on the French dataset (where we top the leaderboard) and on the English-Scientific dataset.
Both these studies show similar trends. As seen in tables \ref{ablation1} and \ref{ablation2}, removing augmented loss i.e. max and mask loss causes a significant drop in the scores, and so does removing the character embeddings. Not including domain-specific language modelling checkpoints leads to drops in scores too, as expected.

\section{Conclusion and Future Work}
In this paper, we introduce the CABACE framework for acronym and long-form extraction that integrates character-level information with mBERT representations and uses domain-specific language modelling and an augmented loss function, with pseudo labelling and adversarial data generation for improved generalizability. Experimental results establish the supremacy of our framework over several baselines on 6 datasets spanning 5 languages. We also evaluate zero-shot cross-lingual efficacy of our proposed model and find that it outperforms baseline mBERT results in 3 out of 4 cases. Our system merits the top spot in French, second place in Danish and Vietnamese and third place in Legal English leaderboards on the AE shared task at SDU AAAI-22.

\begin{figure}[!t]
    \centering
    \includegraphics[width=\linewidth]{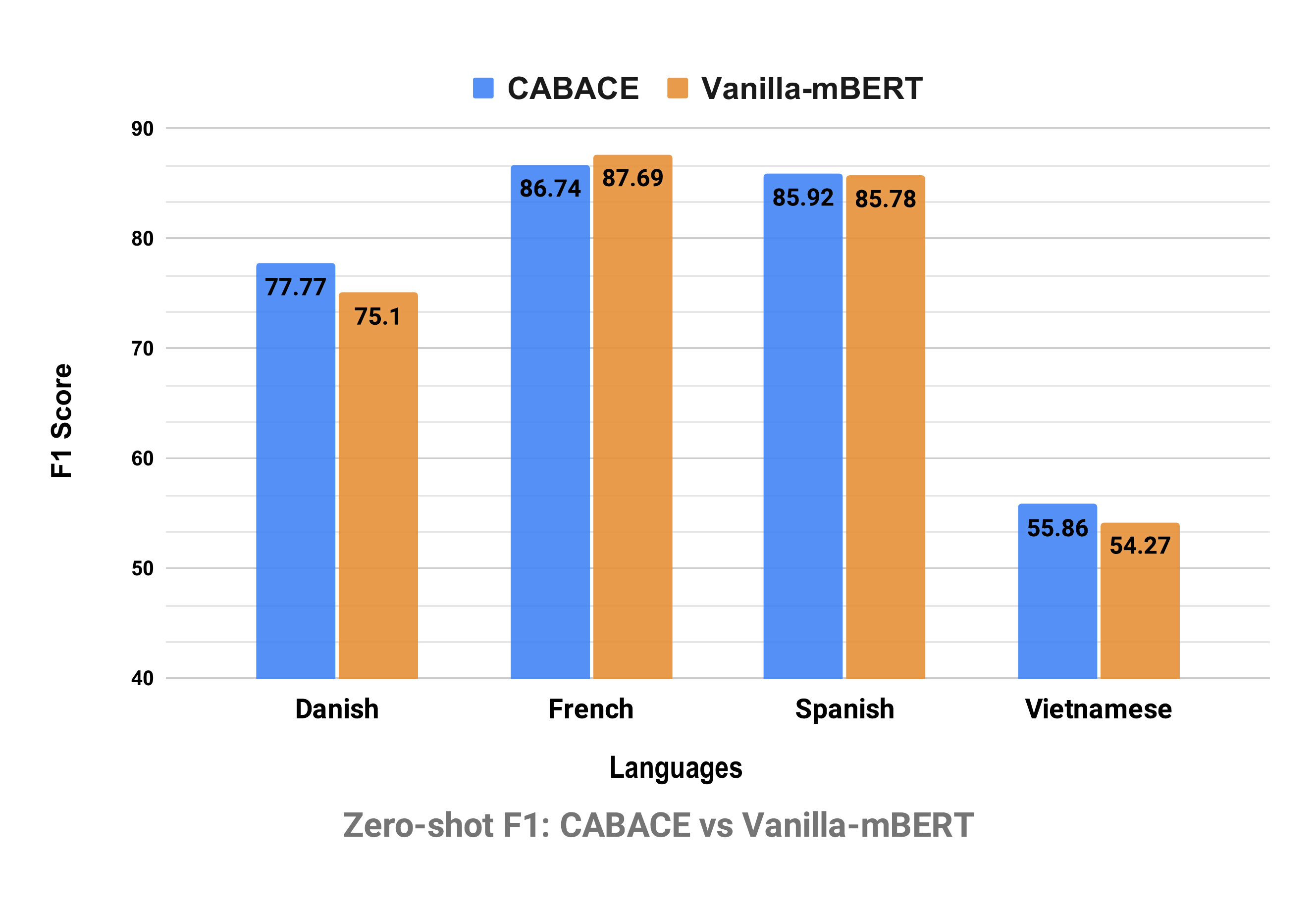}
    \caption{ \footnotesize This figure depicts the zero-shot performance of CABACE as compared to Vanilla mBERT on the non-English languages, with training being done on a combined English-Legal + English-Scientific dataset.}
    \label{fig:zero_shot}
\end{figure}

\begin{table}[!t]
\centering
\small
\begingroup
\setlength{\tabcolsep}{6pt} 
\renewcommand{\arraystretch}{1.2} 

\begin{tabular}{ccccc}
\hline

\textbf{Ablations} & \textbf{Precision} & \textbf{Recall}& \textbf{F1}  \\ \hhline{====}

\multirow{1}{10em}{Ours}  & 0.9387 & 0.9423 & 0.9405 \\ \hline

\multirow{1}{10em}{Ours w/o max loss} & 0.9379 & 0.9426 & 0.9402 \\ \hline

\multirow{1}{10em}{Ours w/o mask loss} & 0.9350 & 0.9403 & 0.9376 \\ \hline

\multirow{1}{10em}{Ours w/o Char Embd} & 0.9361 & 0.9417 & 0.9389 \\ \hline

\multirow{1}{10em}{Ours w/o LM} & 0.9373 & 0.9432 & 0.9402 \\ \hline

\end{tabular}
\endgroup
\caption{CABACE ablation study on French}
\label{ablation1}
\end{table}

\begin{table}[!t]
\centering
\small
\begingroup
\setlength{\tabcolsep}{6pt} 
\renewcommand{\arraystretch}{1.2} 
\begin{tabular}{cccc}
\hline

\textbf{Ablations} & \textbf{Precision} & \textbf{Recall}& \textbf{F1}  \\ \hhline{====}

\multirow{1}{10em}{Ours}  & 0.8282 & 0.8876 & 0.8509 \\ \hline

\multirow{1}{10em}{Ours w/o max loss} & 0.8223 & 0.8811 & 0.8450 \\ \hline

\multirow{1}{10em}{Ours w/o mask loss} & 0.8330 & 0.8822 & 0.8503 \\ \hline

\multirow{1}{10em}{Ours w/o Char Embd} & 0.8254 & 0.8729 & 0.8485 \\ \hline

\multirow{1}{10em}{Ours w/o LM} & 0.8073 & 0.8658 & 0.8355 \\ \hline

\end{tabular}
\endgroup
\caption{CABACE ablation study on English-Scientific}
\label{ablation2}
\end{table}


















Future work on acronym extraction can explore model adaptability to other domains and can attempt to capture acronym--long-form interactions better during their extraction. We use the base version of mBERT for all our experiments; larger and specialized models such as RoBERTa \cite{liu2019roberta}, ELECTRA \cite{clark2020electra}, LegalBERT \cite{chalkidis2020legalbert}, etc. can also be tested.










\section{Acknowledgements}
We thank the organizers of the Acronym Extraction shared task at Scientific Document Understanding (SDU) workshop colocated with AAAI-22 and the anonymous reviewers for their invaluable comments and suggestions.

\bibliography{paper}

\end{document}